\pdfoutput=1 %necessary for arxiv

\documentclass[11pt,a4paper]{article}
\usepackage[]{emnlp2021}
\usepackage{times}
\usepackage{latexsym}
\usepackage{booktabs}

\usepackage{paralist}
\usepackage{tikz}
\usetikzlibrary{positioning,fit,calc,shapes.misc, positioning}

\usepackage{amssymb}% http://ctan.org/pkg/amssymb
\usepackage{pifont}% http://ctan.org/pkg/pifont
\newcommand{\cmark}{\ding{51}}%
\newcommand{\xmark}{\ding{55}}%

\usepackage{lettrine}
\newcommand{\GrantNo}{825303}
\newcommand{\ProjectName}{Bergamot}
\newcommand{\ProjectType}{Research and Innovation Action}

\usepackage{microtype}

\title{Not all parameters are born equal: \\ Attention is mostly what you need}

\author{Nikolay Bogoychev \\
  School of Informatics \\
  University of Edinburgh\\
  United Kingdom \\
  \texttt{n.bogoych@ed.ac.uk} }

\date{}

\begin{document}
\maketitle
\begin{abstract}
Transformers are widely used in state-of-the-art machine translation, but the key to their success is still unknown. To gain insight into this, we consider three groups of parameters: embeddings, attention, and Feed-Forward Neural network (FFN) layers. We examine the relative importance of each by performing an ablation study where we initialise them at random and freeze them, so that their weights do not change over the course of the training. Through this, we show that the attention and FFN are equally important and fulfil the same functionality in a model. We show that the decision about whether a component is frozen or allowed to train is at least as important for the final model performance as its number of parameters. At the same time, the number of parameters alone is not indicative of a component's importance. Finally, while the embedding layer is the least essential for machine translation tasks, it is the most important component for language modelling tasks.
\end{abstract}

\section{Introduction}
Bizarrely, we find that a transformer model ~\cite{vaswani_transformer} for machine translation loses just 1.1--2\% BLEU when any one of the three main components (embeddings, attention, and feed-forward) is frozen with randomly initialised parameters. In more extreme cases we see that with just 6\% of trainable parameters, we maintain 72\% of the baseline BLEU.  This suggests that redundancy in the model architecture is able to compensate for poor components. We perform a case study on transformer style models, exploring the built-in redundancy and the limits of what can be achieved when the majority of parameters are frozen after random initialisation. We do this in several different transformer architecture presets for neural machine translation: transformer-big, transformer-base and a transformer-tiny student preset, and one preset for a language modelling task. Our key findings are:

%One of the common ways to improve the performance of Neural Machine Translation (NMT) \citep{bahdanau_nmt, vaswani_transformer} models is to increase the number of model parameters. This is sometimes done by adding extra components to a model such as a language model \citep{deep_shallow}, but more commonly by adding depth to the model such as in the work of \citet{wmt17_uedin}, or by adding width, as is the difference between transformer-base and transformer-big from \citet{vaswani_transformer}. Through an ablation study where we initialise different transformer components, the embedding layer, the transformer and the feed-forward layer, randomly and the freeze them, we show the following:

\begin{itemize}
    \item The majority of the model information is learned by the attention and the Feed-Forward Neural network (FFN), while the embeddings provide complementary but non-essential signals. The attention and the FFN layers are mostly interchangeable.
    \item Even if it is not essential for some components to be learned, their size (number of parameters) is essential to provide extra expressiveness for the components that are learned: if components are initialised as diagonal matrices as opposed to random matrices, they further lose their effectiveness.
    \item The overall size of the model does affect the results: Larger models have more built-in redundancy and are able to retain higher performance with fewer trainable components.
    \item The smaller proportion of trainable parameters the model has, in general, leads to slower convergence in terms of number of epochs, suggesting learning is hampered.
    \item Our findings are task specific: While machine translation is learned in the attention and the FFN layer, this does not necessarily generalise to other tasks. Language modelling is learned in the embeddings layer.
\end{itemize}

Our study sheds light on the inner workings of the transformer neural network architecture and examines in detail the relative contribution of each component towards the final performance.

%Our study can serve as an optimisation guide when searching for smaller, more computationally efficient architectures that are used in production machine translation systems \citep{wngt19-uedin}.

%We show that even if is not essential to have some components be learned, their size is essential to provide extra non-linearity for the components that
%\begin{itemize}
%    \item[\textbf{Q1}] Which components are the most important for the final result in a transformer model: The embeddings, the attention or the feed forward neural network? Which of these parameters need to be learned?
%    \item[\textbf{Q2}] Is the number of parameters a component contains indicative of its importance for the model?
%    \item[\textbf{Q3}] Does the overall model size affect any of the answers to \textbf{Q1} and \textbf{Q2}? 
%    \item[\textbf{Q4}] Do our findings on machine translation generalise to other tasks, such as language modelling?
%\end{itemize}

\section{Methodology}
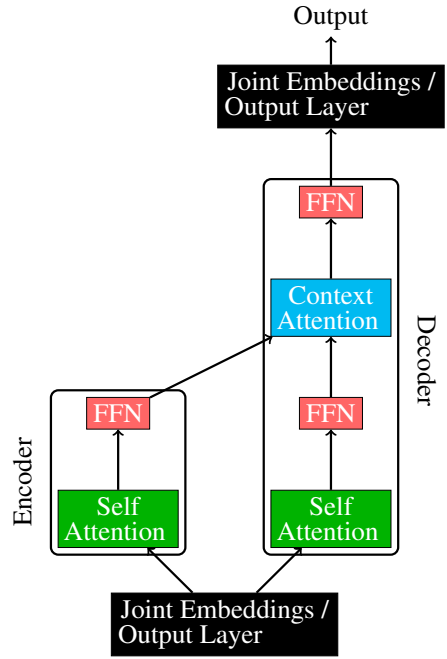
\begin{figure}[ht]
\centering
\scalebox{.7}{
  \begin{tikzpicture}
    \node[draw, fill=red!60,text=white] at (-2,4) (ffnenc) {\centering{\Large{FFN}}};
    \node[draw, text width=2.0cm, fill=black!30!green,text=white] at (-2,2) (selfattenc) {\centering{\Large{Self \\ Attention}}};
    \node[draw, text width=4cm, fill=black,text=white] at (0,0) (emb) {\centering{\Large{Joint Embeddings / \\ Output Layer}}};
    %\node[text width=2cm] at (-1.5,1) (insent) {Input \\ sentence};
    \path [draw=black,->,very thick] (emb) -- (selfattenc);
    \path [draw=black,->,very thick] (selfattenc) -- (ffnenc);
    \node [draw=black, rounded corners, very thick, fit= (selfattenc) (ffnenc) ] (encoder) {};
    \node[rotate=90] at (-3.8,2.7) (encname) {\Large{Encoder}};
    
    \node[draw, fill=red!60,text=white] at (2,8) (ffndec2) {\centering{\Large{FFN}}};
    \node[draw, text width=2.0cm, fill=cyan!80,text=white] at (2,6) (contextdec) {\centering{\Large{Context \\ Attention}}};
    \node[draw, fill=red!60,text=white] at (2,4) (ffndec) {\centering{\Large{FFN}}};
    \node[draw, text width=2.0cm, fill=black!30!green,text=white] at (2,2) (selfattdec) {\centering{\Large{Self \\ Attention}}};

    \path [draw=black,->,very thick] (emb) -- (selfattdec);
    \path [draw=black,->,very thick] (selfattdec) -- (ffndec);
    \path [draw=black,->,very thick] (ffndec) -- (contextdec);
    \path [draw=black,->,very thick] (ffnenc) -- (contextdec);
    \node [draw=black, rounded corners, very thick, fit= (selfattdec) (ffndec) (contextdec) (ffndec2) ] (decoder) {};
    \node[rotate=270] at (3.8,5) (decname) {\Large{Decoder}};
    \node[draw, text width=4cm, fill=black,text=white] at (2,10) (out) {\centering{\Large{Joint Embeddings / \\ Output Layer}}};
    \path [draw=black,->,very thick] (contextdec) -- (ffndec2);
    \path [draw=black,->,very thick] (ffndec2) -- (out);
    \node[] at (2,11.5) (outtxt) {\Large{Output}};
    \path [draw=black,->,very thick] (out) -- (outtxt);
    
  \end{tikzpicture}
  }
    \caption{A simplified view of transformer with input, output embeddings, and output layer tied together. The encoder self-attends to the entire input, while the decoder self-attends to the output it has produced thus far.}
    \label{simpl_trans}
\end{figure}

We perform a parameter-freezing case study on transformer style models. A transformer consists of three basic components: an embeddings layer, attention layers, and FFN, as shown in Figure~\ref{simpl_trans}. The embeddings encompass the source, target embeddings and the output layer as they are tied \citep{tied_embeddings}.

We initialise the model with the Glorot distribution \citep{glorot} and pick a component (or components) to mark as untrainable or ``frozen''.\footnote{Code available at \url{https://github.com/XapaJIaMnu/marian-dev/tree/freeze_FFN}} This effectively means that during training the model will benefit from the extra transformations that those parameters provide, but they are random. It is up to the remaining components to accommodate them. Freezing parameters is frequently performed in transfer learning \citep{zoph-etal-2016-transfer, freezing_nmt_transfer, bert_nmt_frozenemb} in order to prevent the child model from deviating in an undesired manner from the parent. Through these experiments we identify which components are essential for the model, which are interchangeable and which are complementary. An essential trainable component by itself can achieve high performance with the other components being random, while a non-essential component is not able to achieve good performance by itself. Finally, we exchange the frozen random initialisation of components with frozen diagonal (identity) initialisation to verify how important the transformations of the frozen components are.

In the next few sections, we first train a baseline model with no frozen components. Then we train 3 models with one frozen component: either the embedding layer, the FFN layer or the attention. Finally we perform experiments where two out of the three components are frozen.

\section{Transformer-big experiments}
\label{sec:tranbig}
For our main experiments we used 1M parallel Turkish-English sentences from the WMT 18 News translation task \citep{wmt18} (200K gold and 800K back-translated \citep{sennrich-2016-BT}). Data was preprocessed and tokenised using Moses \citep{moses} and a shared vocabulary of 36K byte-pair encoding (BPE) tokens \citep{subword_nmt}. For training the model, we used the Marian \citep{mariannmt} implementation of the transformer-big architecture \citep{vaswani_transformer} with tied embeddings and 8 attention heads for a total of 213M parameters. This model, and all subsequent were trained until 10 consecutive stalls on the perplexity of the dev set or 50 consecutive stalls on the BLEU score of the dev set, whichever is sooner. All the BLEU \citep{bleu} scores are reported on the 2016 dev set provided by the shared task organisers. We present our experiments in Table~\ref{tren}. For reader convenience, each system is also referred to by its number in the result analysis.

\begin{table}[ht]
\centering
\small
\begin{tabular}{@{}lc@{\hskip 0.1in}c@{\hskip 0.1in}crc@{}c@{}}
\toprule
 & \multicolumn{3}{c}{Component} & & & Parameter ratio \\ 
 & EMB & ATT & FFN & BLEU  & Epochs & Trainable/All \\
\midrule
(0) &  \cmark   & \cmark   & \cmark   & 24.3 & 19    & 1                     \\
\midrule
\multicolumn{7}{c}{One frozen component}\\ 
\midrule
(1) & \xmark   & \cmark   & \cmark   & 22.6 & 26    & .82                   \\
(2) & \cmark   & \xmark   & \cmark   & 22.3 & 23    & .64                   \\
(3) & \cmark   & \cmark   & \xmark   & 23.2 & 26    & .52                   \\
\midrule
\multicolumn{7}{c}{Multiple frozen components}\\ 
\midrule
(4) & \xmark   & \cmark   & \xmark   & 21.5 & 36    & .35                   \\
(5) & \xmark   & \xmark   & \cmark   & 20.8 & 37    & .47                   \\
(6) & \cmark   & \xmark   & \xmark   & 4.4  & 25    & .17                   \\
\bottomrule
\end{tabular}
\caption{Trainable and random components for Turkish-English transformer-big. Trainable components are marked with \cmark, fixed components with \xmark. BLEU is computed on the dev set.}
\label{tren}
\end{table}

Freezing a subset of the model leads to a tiny increase in the training speed in terms of words processed per second for all models, due to the reduced computational cost when performing the backward pass. However, this speed benefit is negated by the extra epochs required to achieve convergence, to the point that every single model was slower to converge in terms of real time. The increased training time is likely due to the optimiser running into more saddle points, as there are less parameters that can be pushed into a better model direction \citep{saddle1, saddle2}. Freezing components also always leads to reduced performance in terms of BLEU compared to the baseline, which is expected and is also observed when transferring pretrained components \citep{zoph-etal-2016-transfer, bert_nmt_frozenemb, Aji-transfer}.

%In order to answer \textbf{Q1}, we first define what is an important component. An important component in a transformer which is essential to be learned, as leaving it random will hurt the model more than leaving other components random. 
Our results show that when there is one frozen component, all models perform very similarly, with the frozen FFN one (3) being slightly better than the other two. However, when we look at the experiments where we have multiple frozen components, it is evident that the attention and the FFN layer are more essential for the overall model performance: If both of them are initialised randomly and left frozen, the model fails to learn completely (6).

Based on this we can conclude that the embedding layer is the least essential to be learned in the context of the machine translation task and the remaining components can easily learn to work with random embeddings. This result confirms the findings of \citet{Aji-transfer}, who show that transferring the embedding layer during transfer learning is the least important for the overall model performance. In the rest of the transformer-big experiments, we omit the system with just the embedding layer trainable, as the model quality in this setting is very poor.

We also note that a model with trainable embeddings and attention (3) outperforms a model with frozen embeddings (1), despite the FFN and attention being more essential to be learned. This suggests that the embeddings perform a different role from the attention and the FFN in the translation process. Based on the similar performance that the trainable attention and the FFN offer, we hypothesise that they are able to encode the same information and in case that one of them is not trainable, the other one can mostly fulfil its role. Our hypothesis is supported by the work of \citet{fflayers_keyvalue}, who show that feed forward transformer layers are key-value memories. On the other hand, trainable embeddings, while least essential for the model performance likely learn complementary, albeit non-essential, information to the attention and the FFN.

We find a weak correlation between the number of parameters of a component and its importance to the model: The embeddings layer has the least amount of parameters and is the least essential to be trained, however we can't rank the other two components: The FFN has considerably more parameters than the attention, but they both seem to be equally important for the model's performance.

We note that the model with just attention trainable (4) is able to reach competitive performance, despite having just 35\% of the model parameters.

\subsection{Reducing the embedding layer}
Since we found that the embedding layer is the least essential to be trained, we decided to shrink its dimension from the 1024 (default for transformer-big) to 128, set it to trainable and vary the attention layer and FFN layer. By doing this, we implicitly reduce the width of the other components too, because the matrix dimensions need to match. The resulting models have only 18M parameters (Table~\ref{tren_smallemb}) are markedly worse in terms of BLEU compared to the models with a bigger embedding layer.

\begin{table}[ht]
\centering
\small
\begin{tabular}{@{}lc@{\hskip 0.1in}c@{\hskip 0.1in}crc@{}c@{}}
\toprule
& \multicolumn{3}{c}{Component} & & & Parameter ratio \\ 
& EMB & ATT & FFN & BLEU  & Epochs & Trainable/All \\
\midrule
& \multicolumn{3}{c}{Large baseline}   & 24.3  & 19     & 1                   \\
(0) & \cmark &   \cmark   & \cmark                    & 22.9  & 115    & 1                     \\
\midrule
\multicolumn{7}{c}{One frozen component}\\ 
\midrule
(1) & \xmark &   \cmark   & \cmark                    & 18.9  & 142    & .75                     \\ %model_fixed_emb_small_emb
(2) & \cmark &   \xmark   & \cmark                    & 16.2  & 274    & .93                   \\
(3) & \cmark &   \cmark   & \xmark                    & 18.9  & 132    & .31                   \\
\midrule
\multicolumn{7}{c}{Multiple frozen components}\\ 
\midrule
(4) & \xmark   & \cmark   & \xmark   & 5.1     & 227    & .07                                          \\
(5) & \xmark   & \xmark   & \cmark   & 7.8     & 307    & .69                                  \\
\bottomrule
\end{tabular}
\caption{Turkish-English transformer-big with small, embedding layer dimension. Trainable components are marked with \cmark, fixed components are marked with \xmark. BLEU is computed on the dev set.} %The first row corresponds to the baseline system from Table~\ref{tren}}
\label{tren_smallemb}
\end{table}

These experiments show larger difference in BLEU between attention and FFN: With this smaller model, when embeddings are trainable, a trainable attention (3) reaches much better BLEU score, and in half of the epochs, compared to trainable FFN (2). In addition to that, the attention layer has a lot fewer parameters compared to FFN, showing that the number of parameters a component has is not a marker for how important it is to have that component trainable for the final model's performance. As the difference between the performance of trainable attention vs trainable FFN is more pronounced in the smaller model setting, we can say that model size can change the conclusions. We also note that in this setting, trainable attention and FFN layer (1) achieve the same performance as (3), despite having more than twice the number of parameters, suggesting again that the embeddings are able to learn complementary information to the attention, even if their dimension is tiny.

Furthermore we note that despite that these models are 12 times smaller then the baseline, they converge slower, requiring up to 10 times more epochs, suggesting learning is more difficult. This further highlights that the attention is a more essential component than the FFN, as it can adapt to the training data with fewer signals.

This model has catastrophic loss of BLEU when there is only one trainable component (4, 5), suggesting that larger models are more robust and have more redundancy between components.

\subsection{Reducing the FFN layer size}
We perform a similar experiment with the FFN layer, reducing its dimension from 4096 to 1024. The resulting models have 137M parameters and are shown in Table~\ref{tren_smallattn}.

\begin{table}[ht]
\centering
\small
\begin{tabular}{@{}lc@{\hskip 0.1in}c@{\hskip 0.1in}crc@{}c@{}}
\toprule
& \multicolumn{3}{c}{Component} & & & Parameter ratio \\ 
& EMB & ATT & FFN & BLEU  & Epochs & Trainable/All \\
\midrule
& \multicolumn{3}{c}{trans-big baseline}    & 24.3 & 19    & 1                     \\
(0) & \cmark   & \cmark   & \cmark   & 23.2     & 22    & 1                     \\
\midrule
\multicolumn{7}{c}{One frozen component}\\ 
\midrule
(1) & \xmark   & \cmark   & \cmark   & 22.0     & 38    & .73                   \\
(2) & \cmark   & \xmark   & \cmark   & 20.1     & 36    & .55                   \\
(3) & \cmark   & \cmark   & \xmark   & 23.0     & 18    & .82                   \\
\midrule
\multicolumn{7}{c}{Multiple frozen components}\\ 
\midrule
(4) & \xmark   & \cmark   & \xmark   & 21.6    & 34    & .45                   \\
(5) & \xmark   & \xmark   & \cmark   & 18.0    & 98    & .18                   \\ %startiger likelly stalled
\bottomrule
\end{tabular}
\caption{Trainable and random components for Turkish-English transformer-big with small FFN layer. Trainable components are marked with \cmark, fixed components with \xmark. BLEU is computed on the dev set.} %The first row corresponds to the baseline system from Table~\ref{tren}}
\label{tren_smallffn}
\end{table}

This experiment shows similar trends as the one in the previous section: The model requires more epochs to converge compared to the baseline transformer-big and reaches slightly lower quality. As the size of the FFN layer is reduced, we observe that a trainable attention is always better than a trainable FFN layer (Table~\ref{tren_smallffn}, (2) vs (3), and (4) vs (5)). Finally, when the FFN layer is reduced in size, it matters less whether it is trainable or random and frozen: When it is random (3), we only lose .2 BLEU compared to the baseline system (0). Nevertheless, just the small trainable FFN layer which makes up for 18\% of the total model parameters is able to reach 18.0 BLEU (5).

We note that when all of the components of this model are trainable, it achieves the same performance as the baseline system with frozen FFN (Table~\ref{tren}, (3)). This suggests that for certain components, their width may be as important as whether they are trainable or not. In this setting we also see a sharper drop in performance when attention is not trainable, suggesting the capacity of the FFN to take over the attention's function has diminished.

\subsection{Reducing the attention layer size}
We perform an analogous experiment, where we reduce the size of the second dimension of the keys and queries attention matrices 8-fold, reducing the total model size to 147M parameters. In this setting, all attention matrices make up just 6\% of the total model size. We show our experiments in Table~\ref{tren_smallattn}.

\begin{table}[ht]
\centering
\small
\begin{tabular}{@{}lc@{\hskip 0.1in}c@{\hskip 0.1in}crc@{}c@{}}
\toprule
& \multicolumn{3}{c}{Component} & & & Parameter ratio \\ 
& EMB & ATT & FFN & BLEU  & Epochs & Trainable/All \\
\midrule
& \multicolumn{3}{c}{trans-big baseline}    & 24.3 & 19    & 1                     \\
(0) & \cmark   & \cmark   & \cmark   & 23.4     & 13    & 1                     \\
\midrule
\multicolumn{7}{c}{One frozen component}\\ 
\midrule
(1) & \xmark   & \cmark   & \cmark   & 22.1     & 23    & .74                   \\
(2) & \cmark   & \xmark   & \cmark   & 22.1     & 16    & .94                   \\
(3) & \cmark   & \cmark   & \xmark   & 20.4     & 23    & .31                   \\
\midrule
\multicolumn{7}{c}{Multiple frozen components}\\ 
\midrule
(4) & \xmark   & \cmark   & \xmark   & 17.0    & 196  & .06                   \\ %rindr, ongoing
(5) & \xmark   & \xmark   & \cmark   & 21.3    & 33   & .69                   \\
\bottomrule
\end{tabular}
\caption{Trainable and random components for Turkish-English transformer-big with smaller attention matrices. Trainable components are marked with \cmark, fixed components with \xmark. BLEU is computed on the dev set.} %The first row corresponds to the baseline system from Table~\ref{tren}}
\label{tren_smallattn}
\end{table}

In this setting, our findings are mirroring the results from the previous section: When the size of the attention layer is reduced to that extent, we see that the FFN layer is now more essential for the performance of the model: It outperforms the attention layer in all direct comparisons (Table~\ref{tren_smallattn}, (2) vs (3), and (4) vs (5)). We note that despite the attention layer having only 6\% of the total number of parameters, by itself (4), it manages to retain 72\% of the baseline BLEU score.

We note that in this experiment, as well as in the previous where we reduce the dimension of the FFN layer, there is a clear trend that the number of epochs necessary for convergence increases as the number of trainable model parameters decreases.

\subsection{Self and context attention}
The experiments that we have conducted so far suggest that the attention layer is the most essential for the translation performance. However, attention refers to several different attention modules in a transformer. Attention can be split into self-attention and context attention. Self-attention is present in both the encoder and the decoder. The encoder attends to the full input sequence, while the decoder has masked attention and attends to the output it has produced thus far. Context attention is an attention module that appears only in the decoder and  attends to the encoded sequence. 

We can also make the split by component, having encoder-only attention and decoder-only attention. The different types of attention and their position in the transformer architecture are shown in Figure~\ref{simpl_trans}.

In this experiment, we take the transformer-big model used previously, set the embedding layer to trainable and freeze the FFN layer. We specifically avoid training the FFN layer, because it can take over some of the functionality of our restricted attention layer and this will make our findings less clear. For our experiments, we only train a subset of the attention (the remaining attention parts are left frozen and random), as shown in Table~\ref{attention}.

\begin{table}[ht]
\centering
\small
\begin{tabular}{@{}lcrc@{}c@{}}
\toprule
& & & & Parameter ratio \\ 
& Attention & BLEU  & Epochs & Trainable/All \\
\midrule
(0) & full           & 23.2  & 26     & .52 \\
\midrule
(1) & Self only      & 10.3  & 28     & .41                   \\
(2) & Context only   & 18.1  & 40     & .29                   \\
(3) & Encoder only   & 18.1  & 41     & .29                   \\
(4) & Decoder only   & 19.8  & 24     & .41                   \\
\bottomrule
\end{tabular}
\caption{tr-en transformer-big with trainable embeddings, fixed FFN and a trainable subset of attention. BLEU is computed on the dev set.}
\label{attention}
\end{table}

We see significant drop in BLEU when having only subsection of the attention as learnable. Referring back to our initial experiments in Table~\ref{tren}, partially setting the attention to untrainable is more harmful for the model than freezing it completely and having the FFN trainable.

\citet{lena-heads} show that context heads are the most important heads for a model and are the ones that are pruned last when performing head pruning. In our setting however, even if context heads are not learned, the model can still use the random activations with which they are initialised. That is enough for a model with trainable encoder attention to produce a BLEU score similar to that of a model that uses trainable context attention only. %We have no explanation why trainable self-attention only is outperformed by trainable encoder attention only, as the former trainable parameters are a superset of the latter's trainable parameters.
%We suspect that this is purely an artefact of transformers being notoriously difficult to train and sensitive to hyperparameters \citep{alham-transformers} and it is likely possible to find hyperparameters that are able to bring self-attention-only model, at least on par with the encoder-only. 

The best result is when we have the full attention on the decoder side, as the decoder is the most involved component in translation \citep{lena-heads}.

In our experiments, trainable encoder attention performs better than self-attention only, despite the former having a superset of the latter's trainable parameters (Figure~\ref{simpl_trans}). 

Independently from us, \citet{synthattention} observe that the self-attention component is not essential in the neural machine translation task and can be synthesised on demand. They propose several different functions to do so and achieve smaller quality drop than us, however all of their experiments involve a more advanced synthesisation than simply initialising randomly and not training. \citet{fixed_attn} show that it is possible to replace all but one attention heads with a fixed attentive pattern without impact on translation quality.

Further to this line of work, \citet{MLP_Attention} show that attention could be replaced with a simple gated MLP network and still achieve competitive results, albeit they only show that for vision tasks.

\section{Experiments with smaller models}
The transformer-big preset used in the previous experiments is an overly expressive neural network with enough parameters, so that even when preventing a big part of them from being trained, the model likely has enough built-in redundancy to make up for it. In order to avoid this assumption, we performed analogous experiments with a transformer-base model, as well as a knowledge-distilled student transformer model. These experiments aim to reduce the model size in a systematic way in order to see whether our findings in the previous sections are dependent on the overall model size.

\subsection{Experiments with transformer-base}
We perform the same experiments as the ones in Section~\ref{sec:tranbig}, but we change the model configuration to transformer-base. In effect this means that the number of attention heads remains the same, but the embeddings' and the FFN's dimension is halved. The overall size of the attention is also reduced implicitly, as its dimension is also dependent on the embeddings. The resulting model has 62.6M parameters. We present our results in Table~\ref{tren_base}.

\begin{table}[ht]
\centering
\small
\begin{tabular}{@{}lc@{\hskip 0.1in}c@{\hskip 0.1in}crc@{}c@{}}
\toprule
& \multicolumn{3}{c}{Component} & & & Parameter ratio \\ 
& EMB & ATT & FFN & BLEU  & Epochs & Trainable/All \\
\midrule
& \multicolumn{3}{c}{trans-big baseline} & 24.3 & 19     & 1                     \\
\midrule
(0) & \cmark   & \cmark   & \cmark        & 23.9 & 32     & 1                     \\
\midrule
\multicolumn{7}{c}{One frozen component}\\ 
\midrule
(1) &\xmark   & \cmark   & \cmark        & 22.3 & 62     & .71                   \\
(2) &\cmark   & \xmark   & \cmark        & 20.5 & 40     & .70                   \\
(3) &\cmark   & \cmark   & \xmark        & 22.7 & 39     & .60                   \\
\midrule
\multicolumn{7}{c}{Multiple frozen components}\\ 
\midrule
(4) & \xmark   & \cmark   & \xmark        & 20.0   & 148    & .30                   \\
(5) &\xmark   & \xmark   & \cmark         & 17.5 & 76     & .40                   \\
(6) & \cmark   & \xmark   & \xmark        & 3.9  & 110    & .30                   \\
\bottomrule
\end{tabular}
\caption{Trainable and random components for Turkish-English transformer-base. Trainable components are marked with \cmark, fixed components with \xmark. BLEU is computed on the dev set.} %The first row corresponds to the baseline system from Table~\ref{tren}}
\label{tren_base}
\end{table}

In the transformer-base setting we see similar trends as those with transformer-big (Table~\ref{tren}), except the quality drop when making components untrainable is more severe. We see that the difference between the attention layer and the FFN layer ((2) vs (3), (4) vs (5)) in terms of quality is widening, reaffirming that the attention is the most essential component. The FFN also has more parameters than the attention, showing again that the number of parameters a component has is not indicative of how important it is to have it trainable.

The overall trend we observe is that the smaller the model is, the bigger the gap between the attention and the FFN layer. In a transformer-big, there is more built-in redundancy and when a component is marked as untrainable, it is easy for the other components to step up and make up for it, however as the size of individual components decreases, this is no longer possible.

\subsection{Knowledge distilled models}
We performed an analogous experiment with a knowledge-distilled, small student model \citep{teacher-student}. The student translation models are normally too small to learn a good model from natural data, but given enough synthetic data, they can learn the distribution of a good teacher model. For this setting we trained an ensemble of four transformer-big Estonian-English models on gold and synthetic data from WMT 18 \citep{wmt18}. Then, we decoded 142M sentences with the ensemble and used those to train a student model. The student model has 32K shared source and target vocabulary tokens, embedding dimension of 256, 6 encoder layers, 2 decoder layers and 1536 dimensional Feed forward neural network, similar to the student models used in \citet{Bogoychev-wngt20}. Just as in the previous experiment, the source embeddings are tied to the target embeddings and to the output layer. This model has 16.9M parameters. We report BLEU scores on the dev set provided by the WMT 18 news task organisers \citep{wmt18}. We present our experiments in Table~\ref{eten}.

\begin{table}[ht]
\centering
\small
\begin{tabular}{@{}lc@{\hskip 0.1in}c@{\hskip 0.1in}crc@{}c@{}}
\toprule
& \multicolumn{3}{c}{Component} & & & Parameter ratio \\ 
& EMB & ATT & FFN & BLEU  & Epochs & Trainable/All \\
\midrule
(0) & \cmark   & \cmark   & \cmark   & 24.2  & 14   & 1                     \\
\midrule
\multicolumn{7}{c}{One frozen component}\\ 
\midrule
(1) & \xmark   & \cmark   & \cmark   & 20.3 & 15   & .51                   \\
(2) & \cmark   & \xmark   & \cmark   & 21.6  & 6    & .87                   \\
(3) & \cmark   & \cmark   & \xmark   & 21.3 & 4    & .62                   \\
\midrule
\multicolumn{7}{c}{Multiple frozen components}\\ 
\midrule
(4) & \xmark   & \cmark   & \xmark   & 12.5 & 11   & .14                   \\
(5) & \xmark   & \xmark   & \cmark   & 15.2 & 8    & .39                   \\
(6) & \cmark   & \xmark   & \xmark   & 3.3  & 35   & .49                   \\
\bottomrule
\end{tabular}
\caption{Trainable and random components for Estonian-English student transformer. Trainable components are marked with \cmark, fixed components are marked with \xmark. BLEU is computed on the dev set.}
\label{eten}
\end{table}

In this setting, we observe a few differences from the experiments on bigger transformers: Here, the trainable embedding layer in conjunction with either trainable attention (3), or FFN (2) delivers the best results, outperforming the configuration where it is frozen (1). Despite the embedding layer performing the worst by itself, when it is trainable, it is more valuable to the attention and the FFN. This experiment reaffirms the hypothesis that the attention and FFN layer fulfil the same functionality, and when one of them is frozen, the other is mostly able to compensate for it. However the embedding layer provides a different signal. Again, the embedding layer is unable to learn a meaningful translation model, despite containing nearly half of the model's parameters, clearly showing that the number of parameters is not related to how important a feature is. We also see that this student model is a lot more sensitive to the frozen parameters compared to the bigger transformers, with all results being markedly worse than the baseline, especially the ones with multiple frozen components.

The configuration of this student model differs significantly from the other architectures used, and it also behaves differently when freezing parameters. Therefore, the overall size of the model does affect the behaviour with regards to frozen and random parameters.

\section{Language model experiments}
In order to verify whether our findings generalise to other tasks, we performed a language modelling experiment. We trained a transformer language model, using 78M sentences taken from newscrawl 2017, 2018 and 2019. Our model is based on the transformer-base preset \citep{vaswani_transformer} with 32K BPE tokens vocabulary and 38M parameters.  We show our results in Table~\ref{lmexp}, where we report perplexity on a dev set, formed by taking the first 1000 lines from each of the three newscrawl datasets. The devset is disjoint from the training set.

\begin{table}[ht]
\centering
\small
\begin{tabular}{@{}lc@{\hskip 0.1in}c@{\hskip 0.1in}crc@{}c@{}}
\toprule
& \multicolumn{3}{c}{Component} & & & Parameter ratio \\ 
& EMB & ATT & FFN & PPL  & Epochs & Trainable/All \\
\midrule
(0) & \cmark   & \cmark   & \cmark   & 37.4   & 6   & 1                     \\
\midrule
\multicolumn{7}{c}{One frozen component}\\ 
\midrule
(1) & \xmark   & \cmark   & \cmark   & 118.4 & 6   & .53                   \\
(2) & \cmark   & \xmark   & \cmark   & 47.5  & 6   & .81                   \\
(3) & \cmark   & \cmark   & \xmark   & 50.3  & 6   & .64                   \\
\midrule
\multicolumn{7}{c}{Multiple frozen components}\\ 
\midrule
(4) & \xmark   & \cmark   & \xmark   & 209.3 & 6   & .18                   \\
(5) & \xmark   & \xmark   & \cmark   & 157.3 & 6   & .36                   \\
(6) & \cmark   & \xmark   & \xmark   & 131.7 & 6   & .46                   \\
\bottomrule
\end{tabular}
\caption{Trainable and random components for an English language model. Trainable components are marked with \cmark, fixed components are marked with \xmark. Perplexity is computed on the dev set.}
\label{lmexp}
\end{table}

A neural machine translation system is just a conditional language model, so this task is similar to our previous experiments, except that there is no encoder and no context attention. The transformer components in this case behave differently: the embedding layer is the most essential for the final model performance, capable of delivering the lowest perplexity of any system with multiple frozen components ((2) and (3)), and nearly outperforming by itself (6) a combination of trainable attention and FFN (1). Trainable embeddings and trainable either FFN (2) or attention (3) perform very similarly, suggesting again that they fulfil similar functionality in the model. By itself, however, trainable FFN (5) clearly outperforms trainable attention (4) suggesting that it is more important for this task. It is interesting to note that the number of epochs necessary for convergence does not vary among the different configurations used.

A change in the task can lead to a big change in the overall behaviour of components. For this task, the embedding layer loses the majority of its effectiveness if it is not trainable, unlike in the translation experiments. 

\section{Randomness and transformations}
In order to measure how much the transformations provided by the random parameters help the trainable parameters, we performed an experiment where we initialise either the attention or the FFN layer of a transformer-big model to a diagonal matrix, with zeroes everywhere except at the main diagonal. For the attention this means that all three attention matrices in self and context attention are diagonal and just copy over their input to the output, however since attention is interpreted as different heads, it does mean that the individual heads are very sparse and attend only to a single token. In the case of the FFN layer, since it is not square, the upper part of the matrix is diagonal and the lower part is just zeroes.\footnote{We used \textbf{numpy.eye} to initialise the matrices.} We present our results on transformer-big model from Section~\ref{sec:tranbig} in Table~\ref{frozenzero}.

\begin{table}[ht]
\centering
\small
\begin{tabular}{@{}l@{\hskip 0.08in}c@{\hskip 0.1in}c@{\hskip 0.1in}crc@{}c@{}}
\toprule
& \multicolumn{3}{c}{Component} & & & Parameter ratio \\ 
& EMB & ATT & FFN & BLEU  & Epochs & Trainable/All \\
\midrule
(0) & \cmark   & \cmark   & \cmark   & 24.3 & 19    & 1                     \\
\midrule
\multicolumn{7}{c}{Random frozen component}\\ 
\midrule
(2) & \cmark   & \xmark   & \cmark   & 22.3 & 23    & .64                   \\
(3) & \cmark   & \cmark   & \xmark   & 23.2 & 26    & .52                   \\
\midrule
\multicolumn{7}{c}{Diagonal zeroed frozen component}\\ 
\midrule
(2.1) & \cmark   & \xmark   & \cmark   & 19.4 & 24    & .64                   \\
(3.1) &\cmark   & \cmark   & \xmark   & 22.9 & 20    & .52                   \\
\bottomrule
\end{tabular}
\caption{Trainable and random components for Turkish-English transformer-big contrasting randomly initialised parameters with diagonal matrix parameters. Trainable components are marked with \cmark, fixed components with \xmark. BLEU is computed on the dev set.}
\label{frozenzero}
\end{table}

Our results are conflicting. When the attention is diagonal (3), the BLEU score drops significantly from when it is initialised to random (3.1). This suggests that the model indeed makes use of and learns to work with random parameters and the transformations they provide, however it can not adapt to work with diagonal parameter matrices, as they don't provide useful input.

In the case of the FFN layer, we see the opposite results. Making the FFN layer matrices have 1 only across their main diagonal (2.1) seems to marginally improve performance compared to the baseline case (2). Since the FFN layer is bypassed by residual connections which connect the attention to the output layer, we don't lose any information from the attention, and the ones across the main diagonal of the FFN allows for some minimal mixing of the output of the attention. In this case we can clearly see that the random transformations provided by the untrainable FFN are not as important, as the attention can do its job without them.

We conclude that in certain cases the model does benefit from the transformation provided by the random parameters, however this does not generalise to every component of a transformer. We did not perform this experiment with the embedding layer, as it is not square (The shape is (32000,512) or similar) and it is used to encode different vocabulary items. Rendering it diagonal across the main diagonal would leave the majority of the vocabulary items encoded as a 512 zero vector, which is not something that the model can learn from.

\section{Discussion}

We find that depending on the task, different components of a transformer are essential. Language modelling tasks require the embeddings to be trained, while translation can mostly cope without them. A large amount of research has focused on using BERT \citep{devlin2018bert} or ELMo \citep{elmo} pre-trained embeddings in downstream tasks, but as shown here, embedding representation is not as important as embedding width in a translation task. Furthermore, there is evidence that in some downstream tasks, randomised language models perform competitively \citep{lm_frozen_and_random, past_sesamy_Street}. We urge researchers to always evaluate how essential a component is by setting it to random and frozen, before performing transfer learning from pretrained components.

We find that the "width" of a component is at least as important for the final model performance as whether it is trainable or not, but the raw parameter size of a component is not indicative of whether it is essential to have it trained or not. This clearly suggests that when there is a frozen and random component of a model, the rest of the components are able to work around just as if it were an extra activation function. When the attention is rendered completely linear by making it diagonal, we show a more significant drop in the performance compared to when this is done to the FFN layer. This suggests that the transformation that happens in the attention is more important than the transformation that happens in the FFN layer.

We find that the smaller the model is, the more sensitive it is to having some of its components non-trainable. We suspect that transformer-big models have more built-in redundancy into them and adapt better to this situation. Curiously, the less parameters a model has, or the less trainable parameters the model has, the more epochs it takes to train. This is especially visible in the experiments of transformer-base (Table~\ref{tren_base}) vs transformer-big (Table~\ref{tren}) and the models with reduced embedding layer (Table~\ref{tren_smallemb}). These effects are explainable by the works of \citet{localminimasmallnet, widthbettereasierconv, gradconf} which state that smaller networks are more likely to end up into local minimum and are more sensitive to parameter initialisation compared to bigger networks.

%These effects could be explained by the lottery ticket hypothesis \citep{lotteryTicket}: As the models have less trainable parameters, there is less chance that a parameter has a good initialisation, so more iterations of training are necessary. %This could also be explained by the model having more saddle points during the training \citep{saddle1, saddle2}.

Finally we note how similarly the attention and FFN layers behave: When the attention layer is small, the FFN layer is more essential and vice versa. In a setting where they are trainable by themselves, they achieve similar performance, with the attention having slight edge in larger models. If we decrease the size of the embeddings layer, or use a transformer-base preset, the performance gap between the FFN and the attention widens in favour of the attention. This echoes the findings of \citet{fflayers_keyvalue} and suggests that the attention and the FFN layer are interchangeable and more essential for the model's performance, however there is a degree of redundancy between them. %Better performance is achieved with trainable embeddings and either attention layer or FFN layer, than a combination of trainable attention and FFN layer and non-trainable embeddings.

\section{Related work}
Concurrently and independently from us, \citet{google_echo} perform an ablation study to show how gradually setting different components of an encoder-decoder MT system from random and frozen to trainable increases the BLEU scores from 4 to 39 for an English-French model. Unlike \citeauthor{google_echo} we use a transformer model and we show that wider and deeper models behave differently from small and narrow models. Furthermore we show that depending on the task, the model benefits the most from different components.

Our work follows a long line of research that explores the performance of random (Echo state) neural networks in various tasks \citep{echo_state1, echo_state2, echo_state3, echo_state4}.

\citet{languagePriorsRandom,no_train_encoders} explore random encoders for sentences and evaluate their performance in downstream tasks.

\section{Conclusion and Future work} %Future work for when I have space

We performed a study of transformer neural machine translation component freezing that identified the attention and the FFN as more essential to be learned, but also interchangeable. The embedding layer can not perform the translation task by itself, but the information it learns is complementary to either the attention or the FFN layer. Our findings do not necessarily generalise over other tasks: In the language modelling setting, the components behave differently.

Our findings could serve as the basis for research into new transformer machine translation architectures for production \citep{wngt19-uedin, Bogoychev-wngt20}. We have shown that models can retain competitive performance even when the FFN layer, which accounts for up to 48\% of the parameters, is just a randomly initialised Glorot distribution. This opens the door for models where part of the parameters are generated on the fly during decoding, as opposed to stored statically on disk, making it particularly suitable to embedded devices with limited memory.

%As a future work we propose a neural machine translation system with trainable embeddings and trainable attention, but no FFN layer. Instead the FFN layer can be generated on demand during every single operation by saving a random seed in the model. This can lead to up to 48\% reduction in the model size which could be particularly helpful in a memory limited embedded scenario. This setting also drastically reduces the memory traffic and cache contention in every matrix multiplication that involves the FFN layer. As it well known general matrix multiplication (GEMM) is a memory bound problem so this could lead to increased speed.

\section*{Acknowledgments}

We thank the reviewers for their suggestions and comments! We thank Kenneth Heafield, Adam Lopez, Naomi Saphra and the AGORA research group for the discussions and comments that shaped this paper. We thank Maximiliana Benhke for the use of her pretrained models.

\lettrine[image=true, lines=2, findent=1ex, nindent=0ex, loversize=.15]{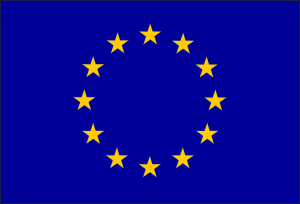}%
{T}his work was conducted within the scope of the \ProjectType\ \textit{\ProjectName}, which has received funding from the European Union's Horizon 2020 research and innovation programme under grant agreement No \GrantNo.

\bibliographystyle{acl_natbib}
\bibliography{emnlp2020}

\clearpage

\appendix
\section{The value of training}
We measured the benefit of training individual components for a short time before fixing them, as opposed to having them untrainable from the beginning. With experiment we show whether training of certain components is more valuable at the beginning of the model training for the final model performance, or they continue to improve and contribute to the final model performance throughout training. We present our results on transformer-big model from Section~\ref{sec:tranbig} in Table~\ref{inittrain}, where we have a model with one frozen component (attention or FFN), where the component is frozen either at initialisation, or after training for few thousand mini-batches.

\begin{table}[ht]
\centering
\small
\begin{tabular}{ccrc}
\toprule
& Frozen after  &  & Total \\
&Epochs & BLEU & Epochs \\
\midrule
& \multicolumn{3}{c}{Pretrained Attention layer}\\ 
\midrule
(1.0) & 0   & 22.3 & 23                \\
(1.1) & 2   & 20.5 & 71                \\
(1.2) & 4   & 20.6 & 71                \\
(1.3) & 7   & 22.2 & 34                \\
\midrule
& \multicolumn{3}{c}{Pretrained FFN layer}\\ 
\midrule
(2.0) & 0   & 23.2 & 26                \\
(2.1) & 2   & 22.6 & 37                \\
(2.2) & 4   & 22.7 & 38                \\
(2.3) & 7   & 22.6 & 21                \\
\bottomrule
\end{tabular}
\caption{Fixed Attention layer or FNN layer for Turkish-English transformer-big contrasting random initialisation with small amount of pretraining. BLEU is computed on the dev set.}
\label{inittrain}
\end{table}

Contrary to our expectations, pretraining the model for a bit before freezing it yields worse overall convergence than having randomly initialised parameters without any pretraining. Pretraining and then freezing either the FFN layer or the attention layer yields similar behaviour. When pretraining is performed for 2 or 4 epochs, we noticed a that the model starts off a bit better than the randomly initialised one, but it quickly catches up in a few epochs. Furthermore convergence time is drastically increased, with small incremental improvements over many more epochs compared to the baseline. 

We note that when the component is pretrained for 7 epochs, the final model performance (and convergence time) starts to approach that of a randomly initialised model, but we do not see any improvements over random initialisation. Looking at the attention, model (1.3) performs similar to the baseline model (1.0), but takes more epochs to converge. When we look at the FFN layer, model (2.3) converges faster than the baseline (2.0), but to a lower BLEU score.

%\section{Appendices}
%placeholder
\end{document}